\documentclass[letterpaper,twocolumn]{article}

\usepackage{fullpage}
\usepackage{setspace}
\usepackage{dblfloatfix}
\usepackage{amsmath}
\usepackage{graphicx}
\usepackage[giveninits=true,sorting=none,style=numeric-comp]{biblatex}
\usepackage{hyperref}

\AtEveryBibitem{%
    \clearfield{urlyear}
    \clearfield{urlmonth}
    \clearlist{language}
}

\bibliography{references}

\DeclareSourcemap{
  \maps[datatype=bibtex]{
    \map{
      \step[fieldsource=doi,final]
      \step[fieldset=url,null]
    }
  }
}

\newcommand{\microsection}[1]{\textbf{#1}}

\title{Agglomerative Attention}
\author{Matthew Spellings\thanks{matthew.p.spellings@gmail.com}}

\begin{document}

\maketitle

\begin{abstract}
Neural networks using transformer-based architectures have recently demonstrated great power and flexibility in modeling sequences of many types. One of the core components of transformer networks is the attention layer, which allows contextual information to be exchanged among sequence elements. While many of the prevalent network structures thus far have utilized full attention---which operates on all pairs of sequence elements---the quadratic scaling of this attention mechanism significantly constrains the size of models that can be trained. In this work, we present an attention model that has only linear requirements in memory and computation time. We show that, despite the simpler attention model, networks using this attention mechanism can attain comparable performance to full attention networks on language modeling tasks.
\end{abstract}

\section*{Introduction}
Recently, neural architectures based on \textit{attention}---rather than recurrence---have driven significant improvements in solving sequence-based problems related to language~\cite{vaswani_attention_2017,dehghani_universal_2018,radford_language_2019,dai_transformer-xl_2019,child_generating_2019}, images~\cite{child_generating_2019}, and even music~\cite{huang_music_2018,payne_musenet_2019}. The core idea of attention is similar in concept to that of learning localized, filter-based features in CNNs~\cite{lecun_gradient-based_1998,krizhevsky_imagenet_2012,zeiler_visualizing_2014,olah_feature_2017}; however, rather than learning \textit{location}-based filters, these attention mechanisms operate on the \textit{content} of sequence elements. For the applications we focus on in this work, full dot-product attention~\cite{vaswani_attention_2017} has been most prevalently used in the literature. This type of attention is quite powerful and has been successfully applied to many types of problems; however, because it operates on all pairs of elements that exist within a sequence (for self-attention) or between target and reference sequences (for general attention), full dot-product attention exhibits quadratic scaling behavior with respect to sequence length. To enable efficient modeling of longer sequences with greater correlation lengths, recent research has focused on finding more scalable attention mechanisms~\cite{shen_bi-directional_2018,child_generating_2019,guo_star-transformer_2019,dai_transformer-xl_2019,ye_segtree_2019}.

In this work we present an attention mechanism that is linear in time and memory requirements. This ``agglomerative attention''---loosely inspired by ideas from protein folding---works by defining a fixed number of classes. Target sequence elements assigned to each class receive a summary representation of all reference elements belonging to that class. We measure the impact of the agglomerative attention algorithm by replacing the full dot product self-attention layers of universal transformers~\cite{dehghani_universal_2018} with agglomerative attention and measure model performance on both character- and word-level language modeling tasks.

\subsection*{Other Attention Models}

\microsection{Full multi-head dot-product attention.} In the typical multi-head dot product attention scheme~\cite{vaswani_attention_2017},  functions are learned that produce key, query, and value vectors $k$, $q$, and $v$ for each pair of elements between two sequences (or, in the case of self-attention, all pairs of locations within a single sequence). From these vectors an attention vector is produced:

\[ \text{attention} = \text{softmax} \left(\frac{q \cdot k}{\sqrt{d}} \right) v \]

\noindent where $d$ is the dimensionality of $k$ and $q$. Typically the attention is split into multiple \textit{heads}, with each head learning its own set of key, query, and value mappings. For applications in self-attention where sequence elements should not be able to ``see the future''---such as language modeling---a mask is applied to prevent queries from attending to keys later in the sequence by setting the softmax argument to a large negative value.

\microsection{Block self-attention networks.} Bidirectional block self-attention networks~\cite{shen_bi-directional_2018} split sequences into fixed-sized blocks. Attention is then applied between all pairs of elements within each block, followed by attention between all pairs of blocks. This enables fine-scale attention for elements within the same block in conjunction with coarser attention between pairs of elements that belong to different blocks.

\microsection{Transformer-XL models.} The Transformer-XL architecture~\cite{dai_transformer-xl_2019} combines attention and recurrence by splitting the inputs into segments. Sequence elements are allowed to attend to elements within the same segment as well as a hidden state from the previous segment, which is propagated between segments much like the hidden state of typical recurrent neural networks. Segmenting the full attention calculation in this way allows the method to transition smoothly between the $O(N)$ runtime of an RNN scheme and the $O(N^2)$ calculations for full attention between all elements as the segment size increases.

\microsection{Star transformers.} Star transformers~\cite{guo_star-transformer_2019} make use of both local and global connections by connecting elements \textit{via} a hub-and-spoke connective scheme. Sequence elements attend directly to their neighbors in the sequence and indirectly to the rest of the sequence through a central \textit{relay node}. This relay node allows information to flow from any node to any other node within two steps of function applications.

\microsection{SegTree transformers.} SegTree transformers~\cite{ye_segtree_2019} perform a binary tree reduction of attention operations over sequence elements, operating in $N\text{log}(N)$ time in the size of the sequence. Depending on the application, the same basic tree structure can be used to summarize (for example, in text classification) or transform (as used in language modeling) the input sequence.

\microsection{Sparse transformers.} Sparse transformers~\cite{child_generating_2019} utilize a sparse attention mask for each sequence element. For example, in an image generation task, one could use a \textit{strided} attention pattern wherein each pixel attends to not only the previous pixel, but also the pixel above or below it. Strided attention is particularly useful for sequences where one can isolate a second dominant lengthscale, as in images~\cite{child_generating_2019} and music~\cite{payne_musenet_2019}.
\section*{Methods}

\subsection*{Agglomerative Attention}

Rather than learning a function to operate on pairs of sequence elements and produce an affinity, \textit{agglomerative attention} groups observations into two or more classes and relays a summary of each class back to the elements of that class. This idea is coarsely inspired by the proposed mechanism of hydrophobic collapse in protein folding~\cite{dill_dominant_1990,agashe_initial_1995,sadqi_how_2003,brylinski_hydrophobic_2006,haran_how_2012} whereby, as one of the first steps of folding, hydrophobic residues rapidly associate with each other in order to minimize exposure to the solvent before forming more refined structural elements. To adapt this biological process into an algorithm for  attention, we choose to classify each sequence element as one of many (rather than two) types and do not concern ourselves with dynamics or details of the interactions between sequence elements, instead simply averaging the representation of all elements for each type.

Given sequences of inputs $x^{\{r,q\}}_{ij}$, with $i \in [1, t^{\{r,q\}}]$ specifying the time indices for reference and query sequences ($r$ and $q$, respectively), and $j \in [1, d]$ specifying an embedding dimension, we first formulate soft class assignments $c_{ik}$ for each sequence element element into one of $k \in [1, m]$ classes:

\[ c^{\{r,q\}}_{ik} = \text{softmax}(x^{\{r,q\}}_{ij} W_{jk}^{\{r,q\}} + b_{k}^{\{r,q\}}) \]

Similar to standard multi-head attention, we keep the representation width constant by projecting the input vector to dimension $d/m$ using a matrix $P^k$ for each class. Projected reference sequence elements are averaged over time using the soft class assignments in one of two ways, depending on whether attending to the future should be allowed (\textit{full}) or not, in the case of self-attention (\textit{masked}):

\[ n_{ik} = \sum\limits_{\tau=1}^{i} c^r_{\tau k} \]

\[a_{i\ell}^{k,\text{masked}} = \frac{1}{n_{ik}} \sum\limits_{\tau=1}^{i}c^r_{\tau k} x_{\tau j} P_{j\ell}^k \]
\[a_{\ell}^{k,\text{full}} = \frac{1}{n_{t^rk}}\sum\limits_{\tau=1}^{t^r} c^r_{\tau k} x_{\tau j} P_{j \ell}^k \]

The averaged projections from each class are then concatenated and multiplied by another matrix to couple the outputs of each class to that of the layer, $q_{ij}$:

\[ p_{ij} = \text{concatenate}(c^q_{i1}a_i^1, c^q_{i2}a_i^2,\ldots, c^q_{im}a_i^m) \]
\[ q_{ij} = Q p_{ij} \]

For self-attention, we maintain separate weights for classifying sequence elements as the reference and query sequence, enforcing a directed information flow from reference to query elements. For ease of reproducibility and comparison with other attention schemes, we integrate agglomerative attention layers into the keras-transformer project~\cite{mavreshko_keras-transformer_2018} as neural network layers using the Keras~\cite{chollet_keras_2015} library.

\section*{Results}

\subsection*{Sequence Length Scaling}

\begin{figure}[h!]
\centering\includegraphics[width=0.48\textwidth]{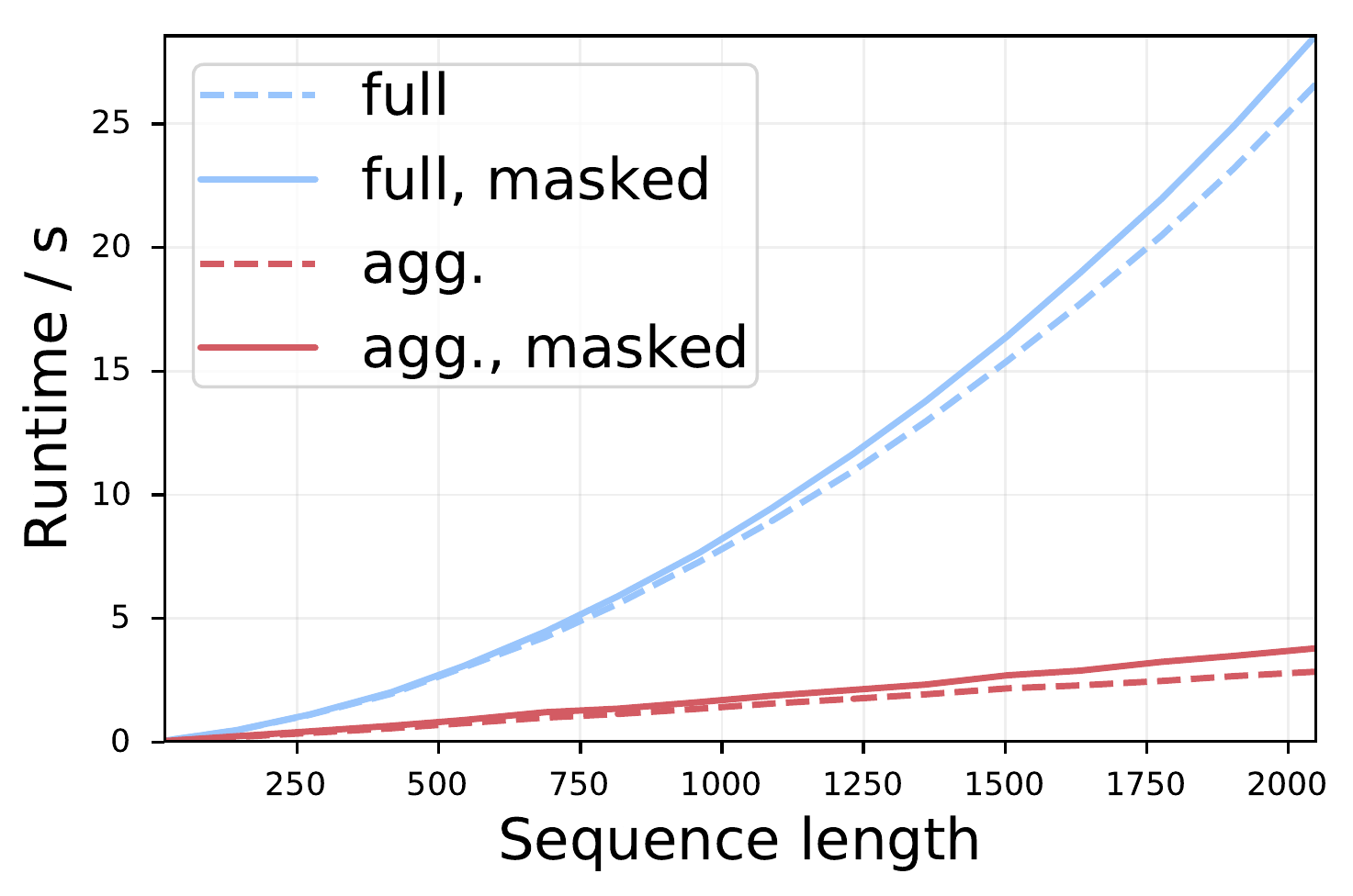}
\caption{
Runtime of individual self-attention layers with respect to sequence size on a single CPU core. Error bars indicating one standard error of the mean over five independent replicas are smaller than the line width and are not included. While full attention exhibits a quadratic runtime with respect to sequence size, agglomerative attention runs in weakly increasing linear time over this range.
}\label{fig:sequence_length_runtime}
\end{figure}

We first analyze the impact of agglomerative attention on computation time for isolated attention layers with realistic embedding sizes and a range of sequence lengths. We measure the time to compute the attention layer output for a single batch of inputs (batch size 32, input dimension 512, 8 heads or agglomerative classes) for both full and agglomerative attention, with and without temporal masking. To better isolate the effect of algorithmic complexity and avoid load balancing and data size issues, we measure the average performance of these algorithms on a single CPU core. Full details of the data generation and analysis are provided in SI Notebook A. Runtime results are presented in Figure~\ref{fig:sequence_length_runtime}.

As can be seen in Figure~\ref{fig:sequence_length_runtime}, agglomerative attention layers quickly begin to outperform full attention with sequence lengths of a few hundred elements. Due to the quadratic scaling of full attention, this gap only widens as the sequence length increases.

\subsection*{Language Modeling Accuracy}

While agglomerative attention is faster, it is also a more simplistic model than dot-product attention. To evaluate the influence of this difference on model accuracy, we perform language modeling using generative pre-training~\cite{radford_improving_2018} of a universal transformer decoder model~\cite{dehghani_universal_2018} on two sets of English-language text: text8~\cite{mahoney_about_2006} and WikiText-2~\cite{merity_pointer_2016}. These models predict the next word or character in the sequence given the text up to that point and the learned weights can be used as a starting point for other tasks, such as machine translation.

To enable a more detailed probe focusing on the impact of agglomerative attention without aggressive optimization of model hyperparameters and learning rate schedules, we train more modestly sized networks (5 transformer blocks with a maximum sequence length of 128 elements and working width of 64 dimensions for character-level modeling and 128 dimensions for word-level modeling) than have been used to obtain state-of-the-art results, using the adadelta optimizer~\cite{zeiler_adadelta_2012}. We halt training after the validation set loss ceases to improve for 10 epochs. In all cases, we report average values over 5 independent replicas, with confidence intervals indicating one standard error of the mean. Full details of this component of the study are provided in SI Notebook B.

Because character-level modeling of text is strongly sequence-dependent---with less meaning associated with individual characters, in contrast to words---we also investigate the use of \textit{causal convolutions}~\cite{bai_empirical_2018} to formulate richer compositions of characters as inputs to the attention layer. Causal convolutions use asymmetric filters that prevent sequence positions from accessing elements that occur later in time. For networks that use convolution-based features, instead of using a sequence embedding as in reference~\cite{vaswani_attention_2017}, we utilize a causal convolution layer with filters of 8 sequence elements. For the sake of completeness, we evaluate the use of convolutions for both character- and word-level language modeling tasks. 
\subsubsection*{Text8: Character-Level Modeling}

\begin{figure}[h!]
\centering\includegraphics[width=0.48\textwidth]{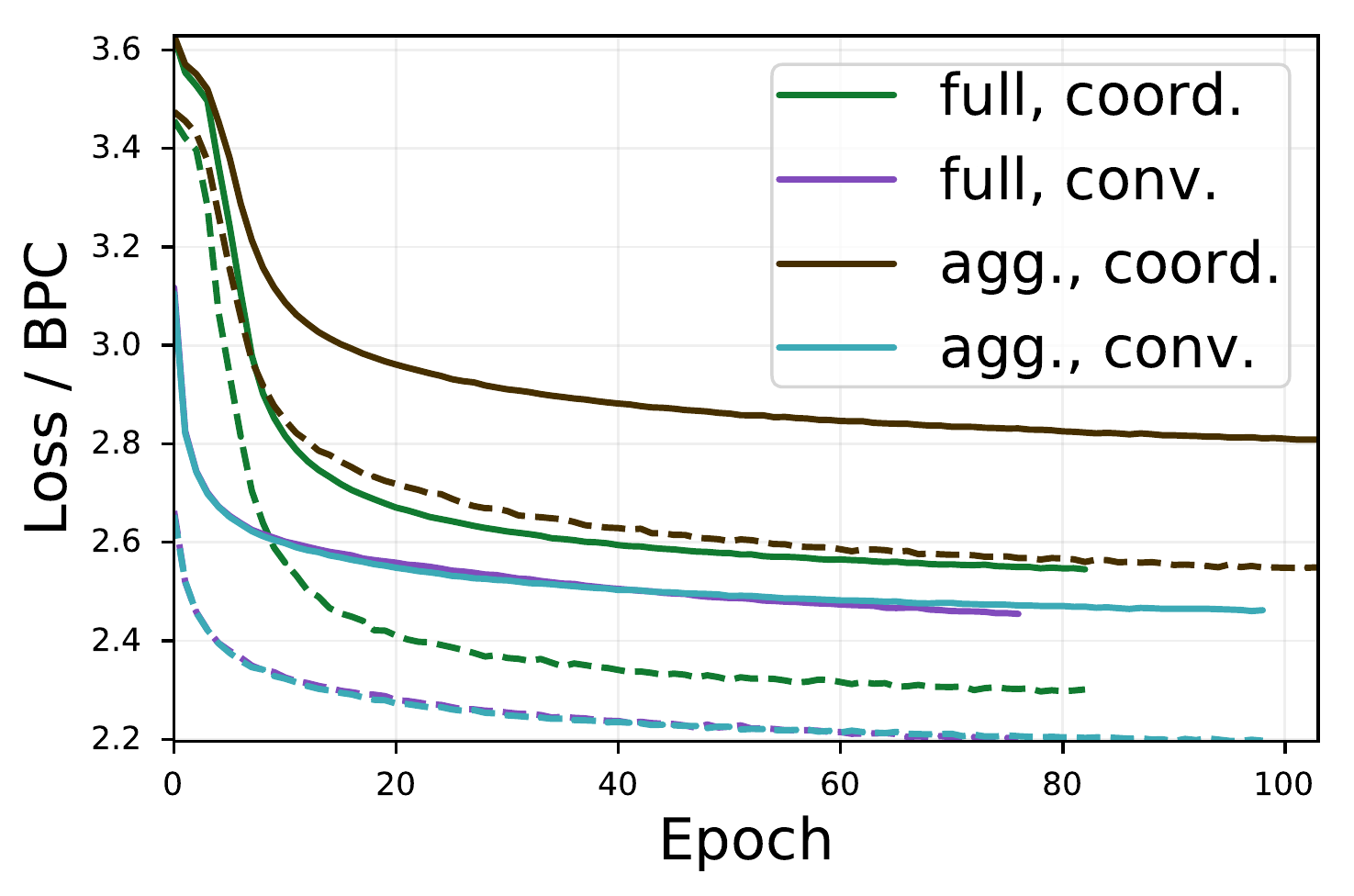}
\caption{
Average training (solid lines) and validation (dashed lines) set losses (in bits per character) of character-level language models trained on the text8 dataset.
Models are trained using full- or agglomerative-style attention, with sequences handled using either sequence embeddings~\cite{vaswani_attention_2017} or causal convolutions~\cite{bai_empirical_2018}.
}\label{fig:text8_training}
\end{figure}

The text8 dataset~\cite{mahoney_about_2006} consists of the first $10^8$ characters of a Wikipedia article dump from 2006. After removing punctuation and markup and converting to lowercase, the remaining contents are spaces and the 26 characters of the English alphabet. We present training curves of language models with both full and agglomerative attention, using position embeddings and causal convolutions, in Figure~\ref{fig:text8_training}.

While agglomerative attention performs significantly worse than full attention when used with position embeddings, convolutions allow the model to learn richer features that yield comparable accuracy to full attention. We find very similar results when evaluating these models on the test set, as shown in Table~\ref{tab:text8_results}.

\begin{table*}[tbp]
\begin{center}
\caption{Number of weights, average test set loss in bits per character (BPC) over five replicas, and typical training time per epoch of character-level models shown in Figure~\ref{fig:text8_training} on the text8 dataset.
}
\begin{tabular}{| c | c | c | c | c |}
\hline
\textbf{Attention type} & \textbf{Sequence encoding} & \textbf{Model size} & \textbf{Test BPC} & \textbf{Epoch time (s)} \\ \hline
Full & Embedding & 64.2K & $2.271 \pm 0.0048$ & 75 \\
Full & Convolution & 88.5K & $2.177 \pm 0.0040$ & 81 \\
Agglomerative & Embedding & 57.0K & $2.52 \pm 0.013$ & 54 \\
Agglomerative & Convolution & 81.4K & $2.183 \pm 0.0035$ & 57 \\ 
\hline
\end{tabular}
\label{tab:text8_results}
\end{center}
\end{table*}

\subsubsection*{WikiText-2: Word-Level Modeling}

\begin{figure*}[h!]
\centering\includegraphics[width=0.96\textwidth]{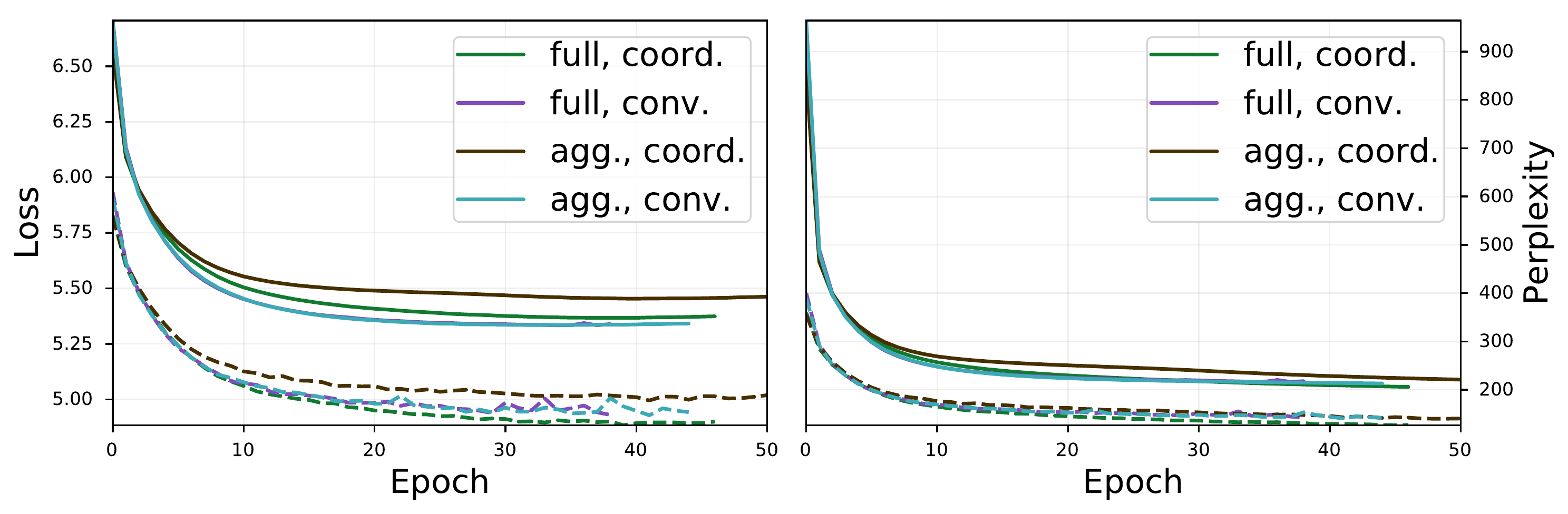}
\caption{
Average training (solid) and validation (dashed) set loss and perplexity of word-level language models for the WikiText-2 dataset over five statistical replicas.
}\label{fig:wikitext2_training}
\end{figure*}

The WikiText-2 dataset~\cite{merity_pointer_2016} contains words from 720 Wikipedia articles with an average length of over 3500 words per article and a total vocabulary of over 33,000 words. Byte pair encoding~\cite{sennrich_neural_2015} is used to formulate a dictionary of word fragments. Training curves for this dataset are shown in Figure~\ref{fig:wikitext2_training}.

While the final relative performance of the models is similar to the results for character-level modeling on the text8 dataset---with the agglomerative attention model utilizing coordinate embeddings performing the worst---the gap between the models narrows much more over time on the WikiText-2 dataset. This is likely due to the increased information content of each word-level sequence element, which makes grouping words \textit{via} convolutions less necessary for this task. Evaluations of these models on test set data are summarized in Table~\ref{tab:wikitext2_results}.

\begin{table*}[tbp]
\begin{center}
\caption{Number of weights, average test set perplexity over five replicas, and training time per epoch of word-level models shown in Figure~\ref{fig:wikitext2_training}.}
\begin{tabular}{| c | c | c | c | c | c |}
\hline
\textbf{Attention type} & \textbf{Sequence encoding} & \textbf{Model size} & \textbf{Test perplexity} & \textbf{Epoch time (s)} \\ \hline
Full & Embedding & 1.50M & $122.0 \pm 0.57$ & 41 \\
Full & Convolution & 1.61M & $134.5 \pm 0.75$ & 44 \\
Agglomerative & Embedding & 1.47M & $134 \pm 1.1$ & 31 \\
Agglomerative & Convolution & 1.58M & $132.6 \pm 0.54$ & 34 \\ 
\hline
\end{tabular}
\label{tab:wikitext2_results}
\end{center}
\end{table*}
\section*{Summary and Conclusion}

Here we presented \textit{agglomerative attention} as a component of transformer neural network architectures to compute attention using linear time and space with respect to sequence length. While we expect agglomerative attention to be less powerful in general than full attention, depending on the model sequence length it could be drastically faster. This increased speed would allow wider networks to be trained within the same computational resource constraints, which can counteract the decrease in accuracy.

Although the scaling-related benefits of this attention mechanism are most well-pronounced in the limit of long sequences, application-specific experiments are necessary to find the optimal number of agglomerative classes. For very long sequences, we may expect narrow networks using few classes to only be able to propagate vague, uninformative summaries of the elements assigned to each class, decreasing the accuracy of the model. For this reason, we recommend that the number of attention classes be optimized as a hyperparameter for practical applications of this method.

We would expect agglomerative attention to be most accurate when sequence elements have rich meaning individually. When the order of elements is of comparable importance to the identity of the element itself---as in character-level modeling of text---creating compositions of elements \textit{via} convolutional layers seems to work well as an alternative.

One could imagine many extensions or variations to the attention model presented here. For example, a hybrid method could involve a coarse, agglomerative-like step taking the top $\ell$ sequence elements for each of $m$ cluster assignments followed by full attention for all elements within each cluster assignment, for work proportional to $m\ell^2$. Similarly, formulating locality-sensitive attention mechanisms---whether locality is directly embedded in the attention layer as in star transformers~\cite{guo_star-transformer_2019} or implicitly included as with the convolutional layer features shown here---seems to be a good method to account for both short-term, detailed correlations at the level of the sequence element as well as longer-term correlations at the sample level.

In summary, we have demonstrated a simple model for attention in sequence-based neural architectures. This simplicity enables better scaling with respect to sequence length at the cost of a less precise attention calculation. We expect the scaling benefits of this method to be most well-pronounced for long sequences, and the attention fidelity to be the best when each sequence element is rich in information content. We hope that models like those presented here will drive far-reaching improvements not only in natural language processing, but also in other places where sequence-based networks have been used by the scientific community, such as chemistry~\cite{schwaller_found_2018} and biology~\cite{rives_biological_2019}.

\subsection*{Acknowledgments}

The author thanks Bradley Dice and Chrisy Xiyu Du for their valuable feedback on this work. The author also thanks Kirill Mavreshko for his work on the keras-transformer library~\cite{mavreshko_keras-transformer_2018}, which provided a convenient foundation for the development of this attention method.

\singlespacing
\printbibliography

\end{document}